\documentclass[11pt]{article}
\usepackage{nodalida2025}
\usepackage{times}
\usepackage{url}
\usepackage{latexsym}
\usepackage{framed}
\usepackage{mdframed}
\usepackage{hyperref}
\usepackage[most]{tcolorbox}
\usepackage{xcolor}
\usepackage{tablefootnote}
\usepackage{bbding}
\usepackage{booktabs}
\usepackage{array}
\usepackage{natbib}

\aclfinalcopy
\title{Optimizing Estonian TV Subtitles with Semi-supervised Learning and LLMs}
\author{Artem Fedorchenko \\
  Tallinn University of Technology \\
  {\tt artem.fedorchenko@taltech.ee} \\\And
  Tanel Alum{\"a}e\\
  Tallinn University of Technology \\
  {\tt tanel.alumae@taltech.ee} \\}
\date{}
\begin{document}
\maketitle
\renewcommand{\abstractname}{\large\bfseries Abstract}  

\begin{abstract}
\small
This paper presents an approach for generating high-quality, same-language subtitles for Estonian TV content. We fine-tune the Whisper model on human-generated Estonian subtitles and enhance it with iterative pseudo-labeling and large language model (LLM) based post-editing. Our experiments demonstrate notable subtitle quality improvement through pseudo-labeling with an unlabeled dataset. We find that applying LLM-based editing at test time enhances subtitle accuracy, while its use during training does not yield further gains. This approach holds promise for creating subtitle quality close to human standard and could be extended to real-time applications.
\end{abstract}

\section{Introduction}
Same-language subtitles for video material, like TV talk shows, investigative pieces, and educational content serve as a valuable resource for the deaf and hard-of-hearing community, non-native speakers and native speakers alike. For instance,  recent studies \cite{preply2024, kim2023comparing} have revealed that 50\% of Americans  and 85\% of the Netflix users overall frequently watch TV and streaming video content with subtitles. Studies show that subtitles can enhance understanding and memory retention. 
A lot of viewers choose to enjoy their content quietly at home, keeping subtitles on to avoid disturbing their roommates or family. 

Subtitles differ from verbatim (word-by-word) transcripts in many aspects. Subtitles represent typically a condensed version of the speech, designed to convey the essential meaning without capturing every word. They may omit filler words, repetitions, and non-verbal sounds, and may rewrite phrases, focusing on clarity and readability for viewers. Since subtitles are displayed on-screen during playback, they are formatted to fit within a limited time frame and limited line length, ensuring they are easy to read while the viewer is watching.

This paper outlines the development of an accurate offline same-language subtitle generation model for Estonian TV content. Using existing human-created subtitles, we fine-tune Whisper \cite{radford2022robust} and explore further improvements with semi-supervised learning and LLM-based post-editing techniques. Our findings demonstrate that Whisper can be trained to closely replicate human subtitling style, creating well-segmented and often rephrased subtitles. Additionally, we find that iterative pseudo-labeling of a large unlabeled dataset improves subtitle quality across all metrics. While a state-of-the-art commercial LLM (OpenAI \textit{gpt-4o}\footnote{We used a regular version of GPT-4o, which was accessed on October 14, 2024.}) can enhance subtitle quality during test time, it's use at training time to improve pseudo-labeled subtitles through post-editing is not effective. 

\section{Related Work}

Both iterative pseudo-labeling and LLM-based post-editing have been an active area of research in the context of verbatim automatic speech recognition (ASR). Pseudo-labeling based semi-supervised learning in ASR has been studied since at least \cite{zavaliagkos1998using} and has been later investigated in several works, e.g. by \citet{vesely2013semi, xu2020iterative}.

To the best of our knowledge, \citet{ma2023can} was the first to show the potential of zero-shot and few-shot LLM-based ASR error correction. This approach has been later extended to take into account uncertainty estimation of ASR outputs \cite{Pu2023MultistageLL} and retrieval-augmented generation for  correcting speech recognition entity name errors \cite{Pusateri2024RetrievalAC}. 

\citet{xi2024semi} showed that LLM-based error correction and data filtering can be also used for refining the pseudo-label transcripts during semi-supervised learning. This work is similar to ours, however, it is applied in the context of a code-switched Mandarin-English ASR task.

\section{Method}

Our method for developing an automated subtitle generation system involves several steps: training with supervised data, using iterative pseudo-labeling, and applying LLM-based error correction.

We start by training the Whisper large-v3 model \cite{radford2022robust} on a supervised dataset. This dataset consists of audio recordings paired with their subtitles.

Next, we use an unsupervised dataset to perform two iterations of pseudo-labeling. In this step, we generate pseudo-labels using the last trained model and combine them with the original supervised dataset, followed by training a new model on this data.

We also apply LLM-based post-editing of the generated subtitles, by instructing the LLM to fix the mistakes in the subtitles and giving it a segment of generated subtitle file. We experiment with applying this LLM-based post-editing in two distinct phases: at test time (i.e., to generated subtitles of the test data) and during training time (i.e., to generated subtitles of the unsupervised dataset).

\section{Experiments}
\subsection{Datasets}

As a supervised dataset\footnote{\url{https://cs.taltech.ee/staff/tanel.alumae/data/etv-subtitles/}}, we used recordings and the corresponding subtitles from the Estonian national TV. The subtitles had been produced for the deaf and hard-of-hearing community by expert subtitlers. The supervised dataset consists of 993 audio-subtitle pairs, totaling 778 hours of audio, corresponding to 10 different TV show series (multi-party talk shows on various topics, political  debates, infotainment programs). We randomly selected 17 recordings out of this set for testing.

The unsupervised dataset contains 7128 audio recordings, amounting to 3923 hours of audio. It contains similar material as the supervised dataset but also contains news program recordings, which the supervised dataset doesn't include.

\subsection{Evaluation metrics}

While evaluating ASR outputs using word error rate (WER) is relatively straightforward, finding an appropriate metric for evaluating automatic subtitling systems is more complicated. Since subtitles often rephrase spoken content to enhance clarity and readability, WER may not accurately reflect the quality of the subtitles. 
WER does also not account for the formatting and timing of subtitles, which are crucial for viewer comprehension. 

In our work, we use three metrics for comparing machine-generated subtitles against reference subtitles: subtitle edit rate (SubER) \cite{wilken2022suber} and two variations of BLEURT \cite{sellam2020bleurt}. SubER is based on a modified version of edit distance that incorporates shifts. This allows it to account for the specific properties of subtitles, such as timing and segmentation. However, SubER doesn't take into account that the same meaning can be conveyed with different words or phrases. 
Thus, we also use BLEURT for evaluation.
BLEURT is a learned metric, trained on subjective human evaluations scores of machine translation  references and the corresponding candidate sentences. BLEURT outputs scores that usually in the range of 0..1 (with 1 being a perfect match) and is found to be better correlated with human judgments in several languages than BLEU scores. We used the multilingual BLEURT-20-D12 model introduced by \citet{pu2021learning}.
Furthermore, we use two variations of BLEURT: t-BLEURT and AS-BLEURT, which differ in the way generated subtitles are aligned to references. AS-BLEURT splits the reference subtitles into sentences, aligns generated subtitles to the references \cite{matusov2005evaluating} and then computes BLEURT score for each sentence, while t-BLEURT does the alignment based on the timing information in the subtitles \cite{cherry2021subtitle}.

\subsection{Baseline Model}
As a baseline model, we finetuned Whisper on our supervised dataset using a cross-entropy objective. The model was trained for 4 epochs using the AdamW optimizer \cite{loshchilov2019decoupled} with a learning rate of \(1 \times 10^{-5}\). We  used an effective batch size of 32 audio chunks  and  applied Stochastic Weight Averaging (SWA) \cite{izmailov2018swa} after the first epoch.

During decoding, we use the Silero VAD model \cite{silero_vad} to remove non-speech parts.

\subsection{Iterative Pseudo-Labeling}

\begin{figure}
    \centering
    \includegraphics[width=\linewidth]{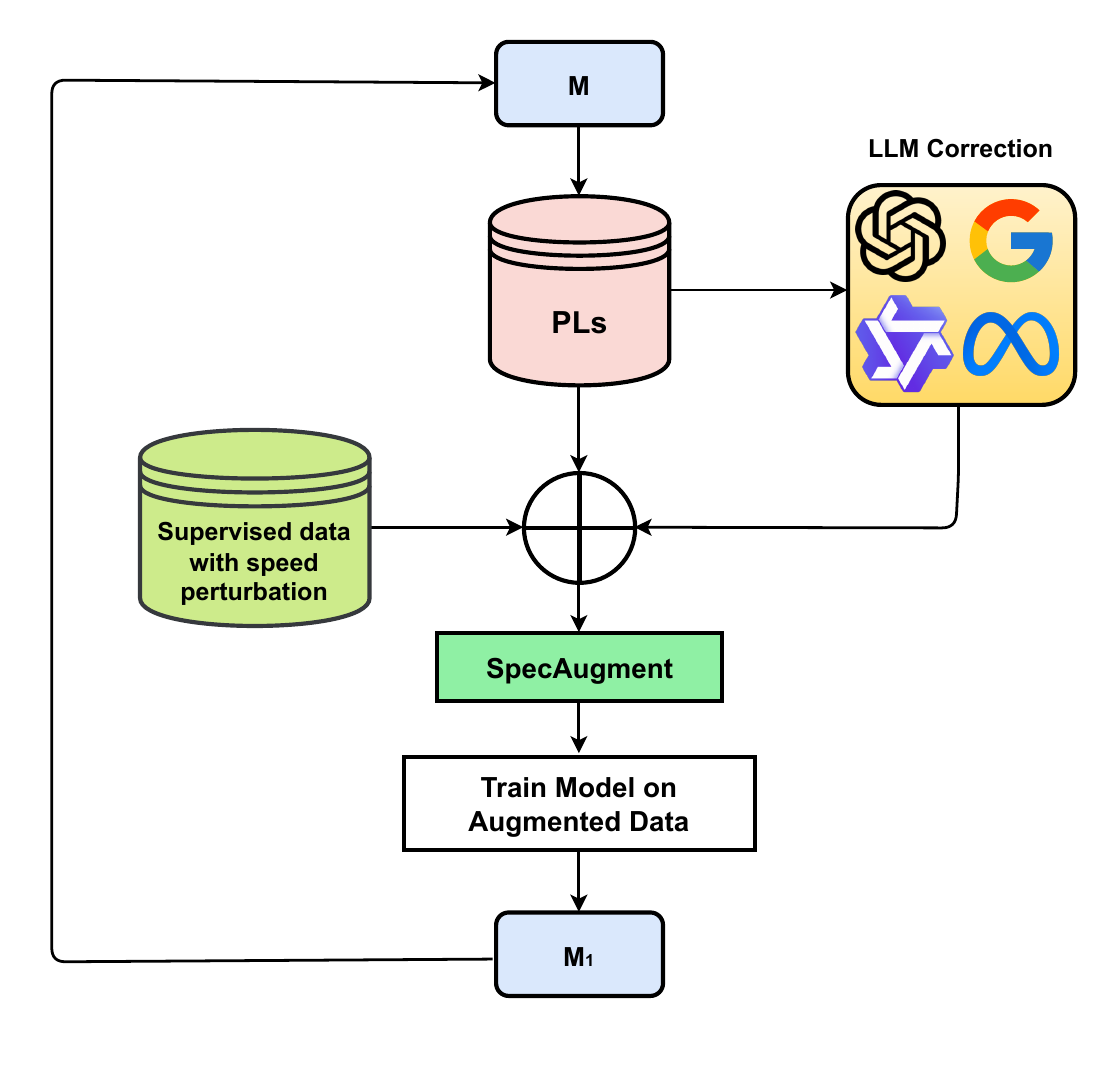}
    \caption{Pseudo-labels generated by model are either passed through LLM or used as as is.}
    \label{fig:NST}
\end{figure}

Next, to improve performance of the baseline model we used \textbf{iterative pseudo-labeling (IPL)} --- a semi-supervised learning technique that enables model refinement on unlabeled data. Starting with an initial model trained on supervised data, we generate pseudo-labels for unlabeled samples and use these to retrain the model iteratively. 

Our approach, which we illustrate in Figure \ref{fig:NST}, explores two strategies for refining pseudo-labels:

\begin{itemize}
    \item \textbf{Direct Pseudo-Labeling}: Using pseudo-labels generated by the model itself.
    \item \textbf{LLM-Enhanced Pseudo-Labeling}: Refining pseudo-labels with a LLM to correct potential errors and ensure alignment with human subtitling standards.
\end{itemize}

In both approaches, we combine pseudo-labeled data with the original supervised dataset, modified by applying speed perturbation. To make the model more robust we applied SpecAugment \cite{Park_2019} on spectrogram level.

We did two iterations of training with pseudo-labels, the training setup was similar to the one with supervised data. 
Additionally, we incorporated weighted loss function:
\[
\mathcal{L}_{\text{total}} = (1 - \lambda) \cdot \mathcal{L}_{\text{supervised}} + \lambda \cdot \mathcal{L}_{\text{pseudo-labels}}
\]

where \(\lambda = 0.35\) was chosen empirically using Optuna \cite{optuna_2019}.

\subsection{LLM-based post-editing}

\begin{figure}[tb]
    \begin{tcolorbox}[colback=gray!10, colframe=black]
        \footnotesize        
        \textbf{System Instruction:} \\[1ex] 
        You are tasked with correcting Estonian subtitles in a subtitle file. \textbf{YOU MUST NOT} create, remove, or modify block numbers and timestamps. \textbf{ONLY} correct the text within the existing blocks.

        \vspace{1em}  
        
        \textbf{Input:} \\[1ex]
        1 \\ 
        00:00:00,000 \texttt{-->} 00:00:02,760 \\ 
        Tere õhtust kõigile, algamas \\ 
        on vestlussaade \textcolor{red}{kahekõne}. 

        2 \\ 
        00:00:02,760 \texttt{-->} 00:00:07,340 \\ 
        Uued rahva poolt palavalt oodatud jõud \\ 
        on \textcolor{red}{toompeal} justkui killustunud. 
        
        \vspace{1em}  
        
        \textbf{LLM Output:} \\[1ex]
        1 \\ 
        00:00:00,000 \texttt{-->} 00:00:02,760 \\ 
        Tere õhtust kõigile, algamas \\ 
        on vestlussaade \textcolor{blue}{"Kahekõne"}. 

        2 \\ 
        00:00:02,760 \texttt{-->} 00:00:07,340 \\ 
        Uued rahva poolt palavalt oodatud jõud \\ 
        on \textcolor{blue}{Toompeal} justkui killustunud. 
    \end{tcolorbox}
    \caption{Example of an LLM instruction used for refining Estonian subtitles. The model corrected the spelling of the TV show name "Kahekõne" and the historical place name "Toompea" in Estonia.}
    \label{fig:llm_prompt}
\end{figure}

To ensure fast and efficient correction of subtitles using an LLM, we split the generated subtitles into chunks of 40 subtitle blocks. This approach allows for great parallelization without exceeding the maximum token limit per request. An example of the request format is shown in Figure \ref{fig:llm_prompt}.

\begin{table}[tb]
\centering
\caption{Comparison of different LLMs for their performance in error correction.}
\label{tab:llms-comparison}
\begin{tabular}{lc}
\hline
LLM            & SubER$\downarrow$ \\
\hline
-              & 35.1 \\
GPT-4o         & \textbf{34.2} \\
Llama 3.1 405B (FP8 quant.)  & 35.5 \\
Qwen 2.5 72B & 36.4 \\
Gemma 2 27B     & 38.4 \\
\hline
\end{tabular}
\end{table}

\newcolumntype{P}[1]{>{\centering\arraybackslash}p{#1}}
\begin{table*}[tb]
\caption{Results of different models, with or without test-time LLM post-editing.}
\label{tab:results}
\footnotesize
\centering
\setlength\extrarowheight{-3pt}
\begin{tabular}{l|P{2.5cm}P{2.5cm}ccc}
\toprule
Finetuning data & Pseudo-label LLM-post-editing? & Test-time LLM-post-editing? & SubER$\downarrow$  & t-BLEURT$\uparrow$  & AS-BLEURT$\uparrow$  \\
\midrule
-                                &     &     & 59.8 & .563 & .728 \\
Verbatim transcripts             &     &     & 51.5 & .526 & .770 \\
\midrule
Subtitles (\textbf{A})                        &     &     & 35.1 & .545 & .799 \\
Subtitles (\textbf{B})                        &     & \Checkmark & 34.2 & .582 & .810 \\
\midrule
\multicolumn{5}{l}{\textit{Pseudo-labeling, iteration 1}} \\
\midrule
Subtitles + pseudo-labels &     &     & 34.5 & .526 & .808 \\
Subtitles + pseudo-labels &     & \Checkmark & 33.9 & .529 & .815 \\
Subtitles + pseudo-labels  & \Checkmark &     & 34.4 & .525 & .810 \\
Subtitles + pseudo-labels & \Checkmark & \Checkmark & 33.9 & .528 & .816 \\
\midrule
\multicolumn{5}{l}{\textit{Pseudo-labeling, iteration 2}} \\
\midrule
Subtitles + pseudo-labels &     &     & 33.4 & .529 & .853 \\
Subtitles + pseudo-labels (\textbf{C}) &     & \Checkmark & \textbf{33.1} & \textbf{.598} & \textbf{.858} \\
Subtitles + pseudo-labels  & \Checkmark &     & 33.6 & .570 & .854 \\
Subtitles + pseudo-labels & \Checkmark & \Checkmark & 33.3 & .571 & .856 \\
\bottomrule
\end{tabular}
\end{table*}

In the development phase, we evaluated several different LLMs for their suitability for this task. Table \ref{tab:llms-comparison} shows the SubER results on test data, after applying LLM-based error correction with different LLMs. We compared OpenAI GPT-4o and three of the best open source LLMs from different vendors. As can be seen, only GPT-4o was able to improve SubER-based subtitle accuracy. 
Based on these results, we used GPT-4o in our experiments.

During our experiments, we observed that LLMs often struggle to output the exact timestamps and block numbers correctly. To address this, we verified these details against the original subtitles to ensure accuracy and re-requested the LLM to fix the issue, if necessary. We also experimented with one-shot and few-shot prompts but did not observe any significant quality improvement, so we opted not to include them. Additionally, we set a threshold on the number of allowable reference check failures: if the model failed more than 3 times, we reverted to the original subtitle.

\subsection{Results}

Table \ref{tab:results} lists evaluation results of the native Whisper model (not fine-tuned on additional data), Whisper fine-tuned on 1066 hours of verbatim transcripts from the TalTech Estonian Speech Dataset 1.0 \cite{alumae-etal-2023-automatic}, and after fine-tuning with different sets of subtitle datasets. The table also highlights the effects of LLM-based post-editing applied during both the training and testing phases.

The results indicate that fine-tuning on subtitle data yields notably lower SubER values compared to fine-tuning on verbatim transcripts, demonstrating the different nature of subtitles and verbatim transcripts. However, the BLEURT scores for both the native Whisper model and the version fine-tuned on verbatim transcripts are surprisingly high. This outcome may be attributed to BLEURT’s design as a semantic similarity metric, which effectively maps both verbatim transcripts and subtitle-like compressed transcripts to proximate points in its semantic space.

To support our interpretation of the achieved results, we computed Wilcoxon signed-rank test \cite{wilcoxon1945individual} between models \textbf{A}, \textbf{B} and \textbf{C} highlighted in the Table \ref{tab:results}. P-value achieved from comparing model \textbf{A} to \textbf{B} is 0.000, \textbf{B} to \textbf{C} is 0.004 and \textbf{A} to \textbf{C} is 0.000. These p-values are all below common significance thresholds (e.g., 0.05), indicating that the differences between the models are statistically significant.

Given that, findings suggest that iterative semi-supervised learning enhances subtitle quality, as evidenced by improvements across all test metrics. LLM-based post-editing applied to decoded subtitles provides additional benefits in most cases. However, contrary to findings in \cite{xi2024semi}, applying LLM-based post-editing to pseudo-labeled subtitles in the unsupervised dataset does not yield further improvements.

Although a formal human evaluation of the generated subtitles was not conducted, the authors' subjective assessment suggests that minimal manual post-editing would be required to achieve error-free subtitles, particularly for in-domain TV data. A sample video from our test dataset, featuring both reference subtitles and subtitles generated by our best model\footnote{\url{https://huggingface.co/TalTechNLP/whisper-large-v3-et-subs}} is available at \url{https://www.youtube.com/watch?v=bEow5vGIgZc}. A smaller version of this model based on Whisper \textit{large-v3-turbo} can be freely used via a simple web application\footnote{\url{https://huggingface.co/spaces/TanelAlumae/whisper-large-v3-et-subs}}.

\section{Conclusion}
In this work, we presented an approach to automated subtitle generation, leveraging the multilingual Whisper model, semi-supervised learning, and LLM-based post-editing. By utilizing supervised and unsupervised datasets, we demonstrated that iterative pseudo-labeling can indeed improve the quality of subtitles. Our results show that applying an LLM during test time has a more significant impact on the results across all the key metrics than during training time. Future work will focus on adapting our approach to real-time scenarios.

\bibliographystyle{acl_natbib}
\bibliography{nodalida2025}
\end{document}